\documentclass[runningheads]{llncs}
\usepackage{graphicx}
\usepackage[utf8]{inputenc}
\usepackage{amssymb}
\usepackage{amsmath}
\usepackage{listings}
\usepackage{caption,rotating}
\usepackage{pdflscape}
\usepackage{afterpage,lscape}
\usepackage{bm}
\usepackage{algorithm}
\usepackage{algpseudocode}
\usepackage{xspace}
\usepackage{color,soul}
\usepackage{booktabs}
\usepackage{hvfloat}
\usepackage{array,graphicx}
\usepackage{booktabs}
\usepackage{pifont}

\newcommand*\rot{\rotatebox{90}}
\newcommand*\rotqb{\rotatebox{30}}

\usepackage{array,multirow,colortbl,graphicx,subcaption}
\usepackage{adjustbox}
\usepackage{url}
\usepackage{kbordermatrix}
\usepackage{slashbox}
\usepackage{color,soul}
\usepackage{array, booktabs, makecell}
\usepackage{siunitx, mhchem}

\newcommand{\bmath}[1]{$\mathbf{{#1}}$}

\newcommand{\CORE}{Core\xspace}
\newcommand{\CROWD}{Cull\xspace}
\newcommand{\SOUL}{{\bf{S}}elective {\bf{Ou}}tlier Ensemb{\bf{l}}es framework (Soul)\xspace}

\newcommand{\myline}{\hline\hline}

\begin{document}
\title{\Large  Graph-based Selective Outlier Ensembles}
%\title{Contribution Title\thanks{Supported by organization x.}}
\author{Hamed Sarvari\inst{1} \and
Carlotta Domeniconi\inst{1} \and
Giovanni Stilo\inst{2}}
\authorrunning{Sarvari et al.}
\institute{George Mason University, Fairfax VA 22030, USA \and
Sapienza, University of Rome
\email{stilo@di.uniroma1.it }\\
%\url{http://www.springer.com/gp/computer-science/lncs} \\
\email{\{hsarvari,cdomenic\}@gmu.edu}}
\maketitle              % typeset the header of the contribution

\begin{abstract} \small\baselineskip=9pt 
An ensemble technique is characterized by the mechanism that generates the components and by the mechanism that combines them. A common way to achieve the consensus is to enable each component to equally participate  in the aggregation process. A problem with this approach is that poor components are likely to negatively affect the quality of the consensus result. To address this issue, alternatives have been explored in the literature to build selective classifier  and cluster ensembles, where only a subset of the components contributes to the computation of the consensus. Of the family of ensemble methods, outlier ensembles are the least studied. Only recently, the selection problem for outlier ensembles has been  discussed.  In this work we define a new graph-based class of ranking selection methods. A method in this class is characterized by two main steps: (1) Mapping the rankings onto a graph structure; and (2) Mining the resulting graph to identify a subset of rankings. We define a specific instance of the graph-based ranking selection class. Specifically, we map the problem of selecting ensemble components onto a mining problem in a graph. An extensive evaluation was conducted on a variety of heterogeneous data and methods. Our empirical results show that our approach outperforms  state-of-the-art  selective outlier ensemble techniques.

\end{abstract}

\section{Introduction}
\label{sec:intro}
An ensemble technique is characterized by the mechanism that generates the components and by the  mechanism that combines them. A common way to achieve the consensus is to enable each component to equally participate  in the process. A problem with this approach is that poor components are likely to negatively affect the quality of the consensus result. To address this issue, alternatives have been explored in the literature to build selective classifier  and cluster ensembles, where only a subset of the components contributes to the computation of the consensus. Typically, in classifier ensembles the selection is driven by the trade-off between accuracy and diversity \cite{Li13}. Boosting, perhaps the most well-known example, achieves the consensus by weighing the components based on their accuracy. In an unsupervised scenario, such as clustering and outlier detection, defining the selection mechanism is more challenging due to the lack of ground truth. Quality and diversity have been used as measure to drive the selection of components for cluster ensembles \cite{Fern08}. 

Of the family of ensemble methods, outlier ensembles are the least studied \cite{zimek2014ensembles,schubert12,rayana2015less,aggarwal15,Aggarwal13,zimek2013subsampling,Nguyen10,Micenkova15,Liu08}.  In particular, only recently the selection problem for outlier ensembles has been  discussed \cite{schubert12,rayana2015less}, and its potential positive effect on event detection has been shown \cite{rayana2015less}. 
In this work, we further explore the selection issue for outlier ensembles, and define a new \textit {graph-based class} of ranking selection methods, of which we detail specific instances.

\begin{figure}[!htbp]
\begin{subfigure}[b]{\columnwidth}
	\includegraphics[width=0.98\columnwidth]{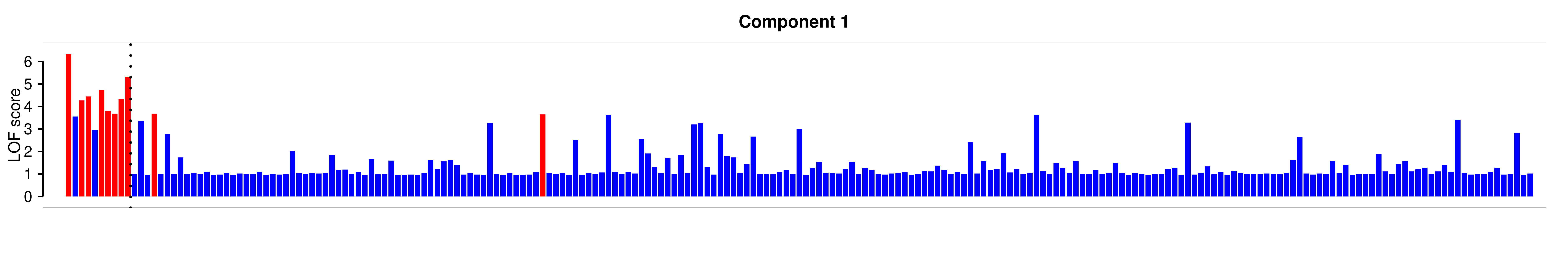}
	\caption{}
	\label{fig:comp_dist:a}
\end{subfigure}  
\begin{subfigure}[b]{\columnwidth}
	\includegraphics[width=0.98\columnwidth]{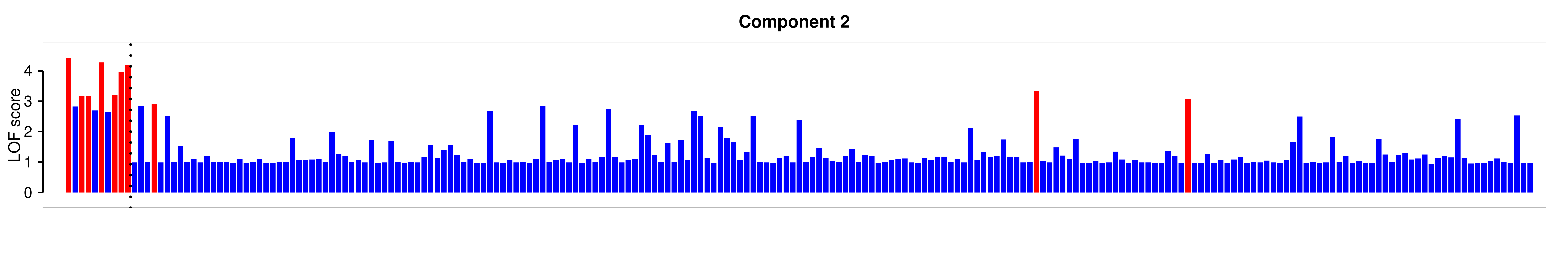}
	\caption{}
	\label{fig:comp_dist:b}
\end{subfigure}  
\begin{subfigure}[b]{\columnwidth}
	\includegraphics[width=0.98\columnwidth]{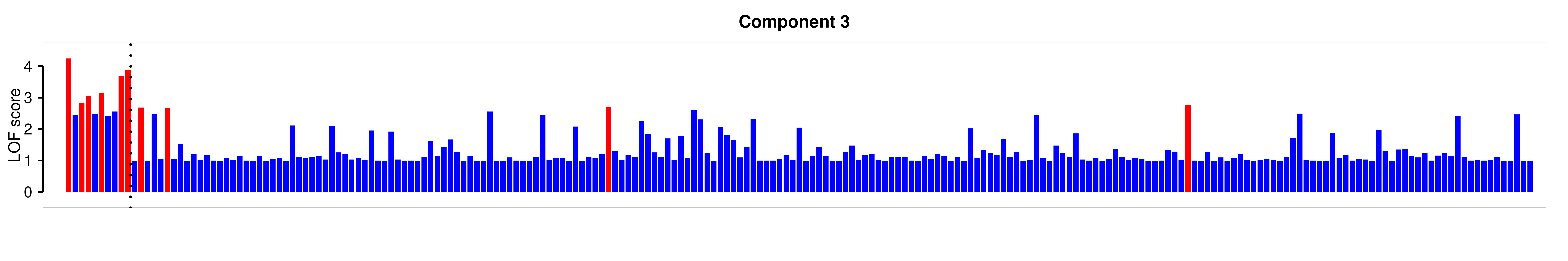}
	\caption{}
	\label{fig:comp_dist:c}
\end{subfigure}  
\begin{subfigure}[b]{\columnwidth}
	\includegraphics[width=0.98\columnwidth]{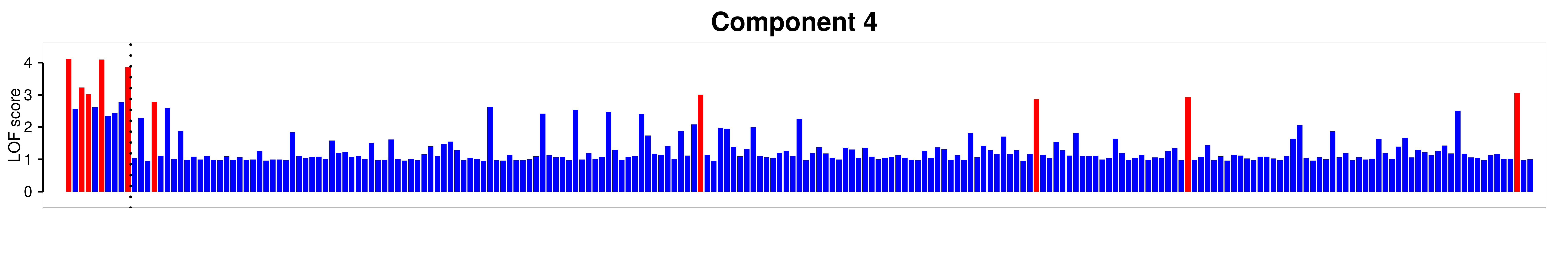}
	\caption{}
	\label{fig:comp_dist:d}
\end{subfigure}  
\begin{subfigure}[b]{\columnwidth}
	\includegraphics[width=0.98\columnwidth]{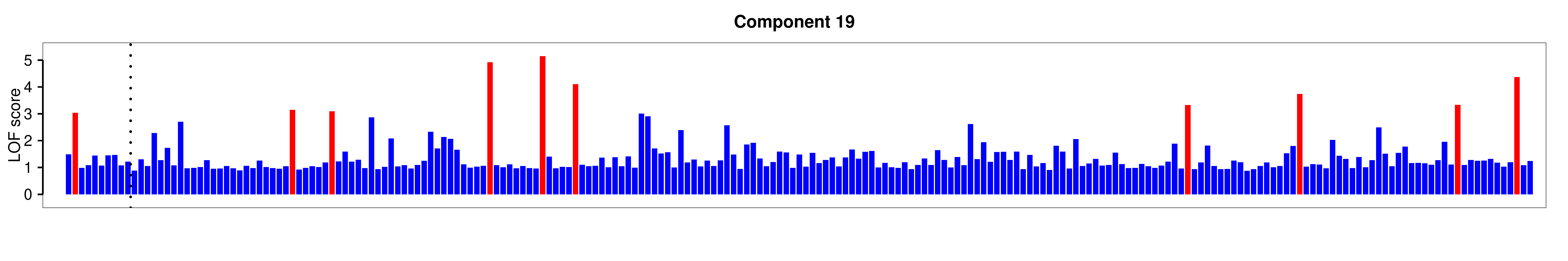}
	\caption{}
	\label{fig:comp_dist:e}
\end{subfigure}  
\begin{subfigure}[b]{\columnwidth}
	\includegraphics[width=0.98\columnwidth]{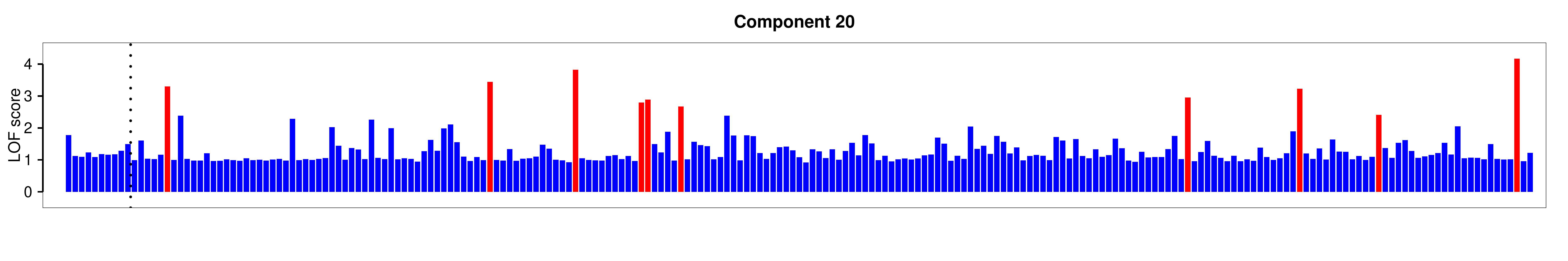}
	\caption{}
	\label{fig:comp_dist:f}
\end{subfigure}  
\begin{subfigure}[b]{\columnwidth}
	\includegraphics[width=0.98\columnwidth]{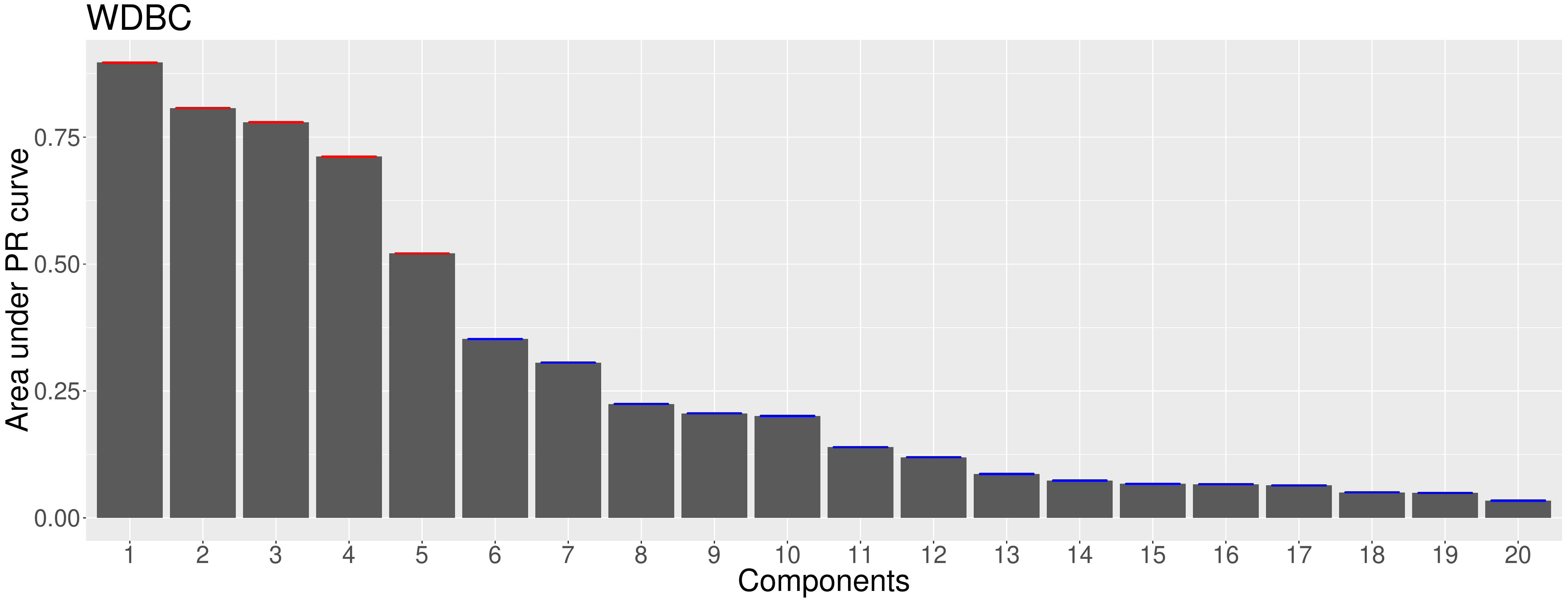}
	\caption{}
	\label{fig:comp_dist:g}
\end{subfigure} 
\caption{WDBC data set. (a)-(f): Outlier scores generated using the LOF algorithm; (g): Area under the PR curve for 20 components. \label{fig:comp_dist}}
\end{figure}

To better understand the nature of the problem we want to tackle, let's consider Figure \ref{fig:comp_dist}. Plots (a)-(f) show six ranking components generated from the WDBC data using the LOF algorithm under different conditions (see Section \ref{sec:exp} for details). Each row corresponds to a ranking. The horizontal axis captures the data points (in a fixed order across all six rows), and the vertical axis measures the LOF scores assigned to each point. The 10 leftmost points are the actual outliers, and the red vertical bars highlight the top-10 LOF score values, in the respective rankings. The four rankings (a)-(d) identify many of the outliers among the top-10 ranked points, while the rankings (e) and (f) have at most one outlier among the top-10 ranked points. Figure \ref{fig:comp_dist}(g) shows the area under the PR curve (AUCPR) for an ensemble of 20 rankings, of which six are the ones illustrated. Rankings (a)-(d) correspond to the most accurate ones, and (e)-(f) are the two least accurate. As also observed in \cite{rayana2015less}, the best rankings tend to agree on the high scored points, but the actual scores change. As a consequence, their aggregation may produce an improved ranking. On the other hand, rankings (e) and (f) largely rank non-outliers as the highest, and including them in the aggregation process may affect the consensus ranking negatively. We aim at identifying such poor rankings and remove them from further consideration. Our technique, called \CORE (described in Section \ref{sec:RS}), was able to select the five top rankings from the ensemble of 20 components given in Figure \ref{fig:comp_dist}.
The consensus ranking achieved via averaging the selected five components gives an AUCPR of 0.8, while the consensus ranking achieved via averaging the entire 20 components gives an AUCPR of 0.2. This result is indicative of the great potential our graph-based approach to selective outlier ensembles has to offer. 

The paper is organized as follows. Section \ref{sec:rel-Works} discusses related work. We introduce our framework and methodology in Sections \ref{sec:soul} and \ref{sec:RS}, and in Section \ref{sec:exp} we present our experiments, results, and analysis. Section \ref{sec:conclusion} concludes the paper with thoughts for future work.

\section{Related Work}
\label{sec:rel-Works}
Ensemble methods have been exploited in the literature to boost the performance of classifiers, 
e.g. \cite{dietterich2000ensemble,breiman1996bagging,freund1995desicion,breiman2001random,domingos2000bayesian,dvzeroski2004combining}, and more recently clustering ensembles have emerged as a technique for overcoming problems with clustering algorithms, e.g. \cite{strehl2002cluster,bickel2004multi,muller2012discovering}. 
It is well known that off-the-shelf clustering methods may discover different patterns in a given set of data. 
This is because each clustering algorithm has its own bias resulting from the optimization of different criteria. Furthermore, there is no ground truth against which the clustering result can be validated. 
Thus, no cross-validation technique can be carried out to tune input parameters involved in the clustering process. 
Clustering ensembles offer a solution to challenges inherent to clustering arising from its ill-posed nature:
they can provide more robust and stable solutions by making use of the consensus across multiple clustering results, while averaging out emergent spurious structures that arise due to the various biases to which each participating algorithm is tuned, or to the variance induced by different data samples.

Anomaly (or outlier) detection is another unsupervised problem that suffers from many of the same challenges as clustering. As such, many different anomaly detection techniques (e.g., density-based and distance-based, global vs. local), and multiple variations of each have been studied \cite{knorr1997,ramaswamy2000,zhang2009,bay2003mining,jin2001mining,lof,loci,jin2006ranking,kriegel2009loop,knox1998algorithms}. A comprehensive survey of these methods can be found in \cite{chandola2007outlier}. Invariably, all anomaly detection algorithms involve parameters that are problematic to set. Ensemble techniques can provide a framework to address  these issues for anomaly detection algorithms, in a way similar to clustering ensembles. Nevertheless, the discussion on outlier ensembles has started only recently \cite{zimek2014ensembles,schubert12,rayana2015less,aggarwal15,Aggarwal13,zimek2013subsampling,Nguyen10,Micenkova15,Liu08}, and the avenue remains largely unexplored. 

In this paper we focus on the problem of component selection for outlier ensembles. As discussed above with the example shown in Figure \ref{fig:comp_dist}, poor components can negatively affect the consensus ranking. Only recently the selection problem for outlier ensembles has been  discussed \cite{schubert12,rayana2015less}, and its potential positive effect on event detection has been shown \cite{rayana2015less}. 
 The selection models presented in \cite{schubert12} (DivE) and in \cite{rayana2015less} (SelectV) are both greedy selective strategies based upon a target ranking treated as  pseudo ground-truth. Rankings are selected in sequence if they increase the weighted Pearson correlation of the current ensemble prediction with the target vector. SelectH \cite{rayana2015less} is also based on a pseudo ground-truth, for anomalies this time. Components that do not rank the estimated anomalies sufficiently high, are candidate to be discarded. To compute the pseudo ground-truth, a mixture modeling approach is used to convert each component ranking into a binary vector. A majority vote applied to these binary lists identifies the anomalies.
 
In this work, we take a different approach. We tackle the problem of selecting outlier ensemble components by mapping it onto a graph mining problem, which does not use the concept of pseudo ground-truth. The main idea is to capture high quality rankings as nodes forming patterns in a graph. We advocate that such transformation can lead to a family of new effective approaches for the selective outlier ensembles problem. The presentation of our framework follows.

\section{Selective Outlier Ensembles Framework}
\label{sec:soul}

 Let $X= \{ {{\bf {x}}_i}  \}_{i=1}^n$, ${{\bf {x}}_i} \in \mathbb{R}^d$, be a collection of data. From the data, a collection of outlier score rankings $\{ r_j\}_{j=1}^m$  is generated. Each ranking is a sequence of $n$ outlier scores, one for each data point, sorted in non-decreasing order: $r_{j1} \geq r_{j2} \geq \dots \geq r_{jn}$. The collection $\{ r_j\}_{j=1}^m$ constitutes the ensemble.
 
In Algorithm \ref{alg:SOul}, the \SOUL is presented. The Soul framework takes in input the collection of rankings $\{ r_j\}_{j=1}^m$ (however generated), and applies a two-phase algorithm.
 The first step is the \textit{Ranking Selection} phase, which allows to plug in {\textit{any}} selection function that specifies which rankings to retain. The \textit{Ranking Aggregation} phase enables the use of a variety of aggregators to compute a consensus ranking $r^{\ast}$. The Soul framework does {\textit{not}} require access to the original features of the data, and is transparent to the process that generates the ensemble. Soul can therefore be used with any outlier detection algorithm that produces a ranking, and any combination thereof.
 
The two steps \textit{Ranking Selection} and \textit{Ranking Aggregation} can also be merged to produce the consensus ranking $r^{\ast}$. In this work, though, we focus on the design of an effective {\textit{Ranking Selection}} function of the components. As such, we make a distinction between the two phases, and apply only commonly used consensus functions, i.e. Maximum, Average, and Minimum \cite{lof,loci,lazarevic2005feature}, for \textit{Ranking Aggregation}.

\begin{algorithm}[H]% Use "stay right HERE" already!
  \begin{algorithmic}[1]
  \Require A collection of rankings $\{ r_j\}_{j=1}^m$. 
  \Ensure A consensus ranking $r^{\ast}$.
  \State {\textit{Ranking Selection}};
  \State {\textit{Ranking Aggregation}};
  \State \Return $r^{\ast}$;
  \end{algorithmic}
  \caption{The {\textit{Soul}} Framework \label{alg:SOul}}
\end{algorithm}
\vspace{-3em}
\begin{algorithm}[H]% Use "stay right HERE" already!
  \begin{algorithmic}[1]
  \Require A collection of rankings $\{ r_j\}_{j=1}^m$. 
  \Ensure A subset of rankings $\{r^{\ast}_j\}_{j=1}^l$.
  \State Construct the complete weighted graph $G_c$;
  \State Derive the pruned graph $G$;
  \State Compute the $k$-core subgraph of $G$ with largest $k$; 
  \State \Return Vertices (rankings) with largest coreness values;
  \end{algorithmic}
  \caption{{\textit{\CORE}} Ranking Selection \label{alg:RS1}}
\end{algorithm}
\vspace{-3em}
\begin{algorithm}[H]% Use "stay right HERE" already!
  \begin{algorithmic}[1]
  \Require A collection of rankings $\{ r_j\}_{j=1}^m$. 
  \Ensure A subset of rankings $\{r^{\ast}_j\}_{j=1}^l$.
  \State Construct the complete weighted graph $G_c$;
  \State Compute weighted degrees of each node;
  \State Discard nodes with lowest weighted degree; 
  \State \Return Remaining vertices (rankings);
  \end{algorithmic}
  \caption{{\textit{\CROWD}} Ranking Selection \label{alg:RS2}}  
\end{algorithm}
%\vspace{-em}

\section{Ranking Selection}
\label{sec:RS}

%%%%%%%%%%%%%%%%%%%%%%%%%%%%%%%%%%%%%%%%%%%%%%%%%
%\begin{figure}[t]
%\centering
%\includegraphics[width=0.4\columnwidth]{Pruned_graph_WBC3-2.pdf}
%\caption{Core ranking selection: Graph $G$ obtained for the 20 rankings of Figure \ref{fig:comp_dist}(g). \label{fig:graph}}
%\end{figure}
\begin{figure}
\begin{subfigure}{0.5\linewidth}
\centering
\includegraphics[width=0.6\columnwidth]{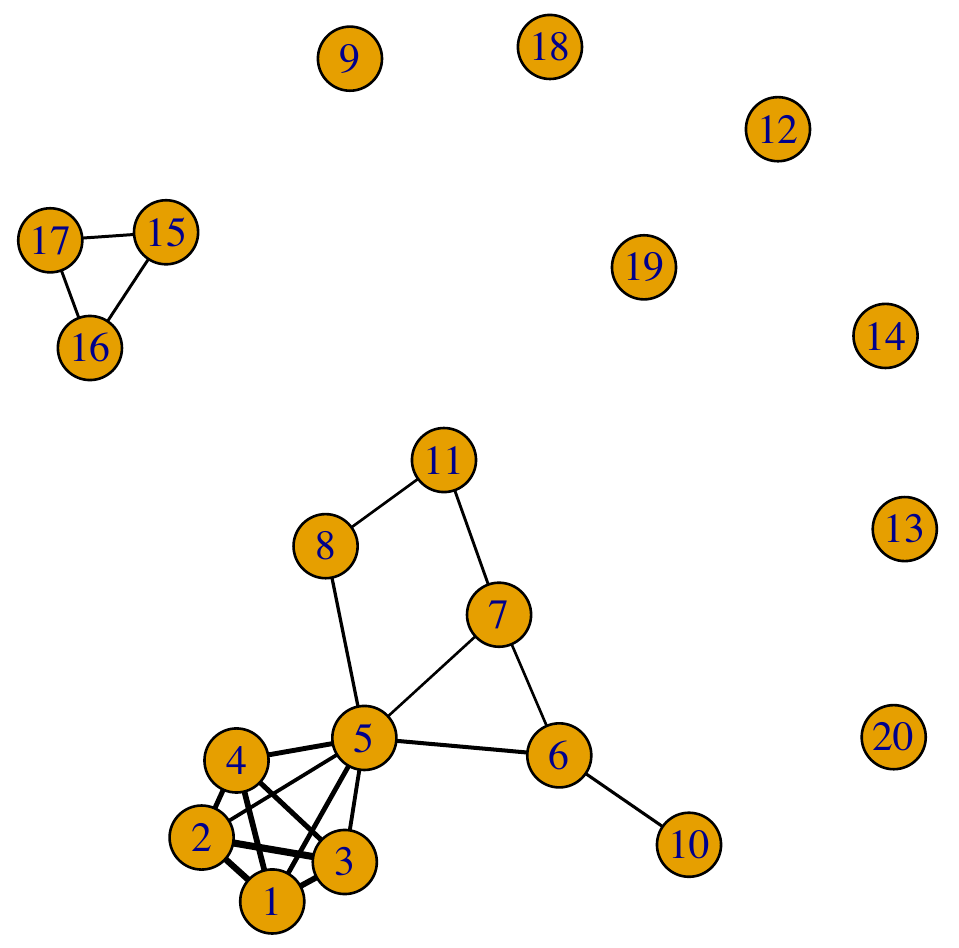}
\caption{\label{fig:graph}}
  \end{subfigure}  
  \begin{subfigure}{0.5\linewidth}
    \centering
    \[
\kbordermatrix{
& \textbf{1} & \textbf{2} & \textbf{3} & \textbf{4} &  \textbf{\dots} & \textbf{19} & \textbf{20}\\
    \textbf{1}&-& 0.92& 0.91 &0.91&  \dots& 0.55 &0.38 \\
	 \textbf{2} &0.92& - &0.93 &0.89 &  \dots&0.52 &0.38 \\
	\textbf{3} & 0.91& 0.93& -& 0.90&  \dots& 0.53& 0.39 \\
	 \textbf{4} &0.91& 0.89& 0.90& -&  \dots& 0.55& 0.41 \\
	\dots & \dots & \dots & \dots & \dots & - & \dots& \dots \\
    \textbf{19} &  0.55& 0.52& 0.53& 0.55&  \dots& -& 0.58 \\
	 \textbf{20}& 0.38& 0.38& 0.39& 0.41&  \dots& 0.58& - 
}
\]
\caption{ \label{tab:wtau}}
  \end{subfigure}   
  \caption{(a) Core ranking selection: Graph $G$ obtained for the 20 rankings of Figure \ref{fig:comp_dist}(g).
  (b) Weighted Kendall tau similarity values between the six rankings given in Figure \ref{fig:comp_dist}.} 
\end{figure}

We define a {\textit{graph-based}} class of ranking selection methods. A method in this class  is characterized by two major steps:
\begin{enumerate}
\item Mapping the rankings onto a graph structure
\item Mining the resulting graph to identify a subset of rankings.
\end{enumerate}
$$ \{r_j\}_{j=1}^m \xrightarrow[\text{}]{\text{mapping to a graph}}G\xrightarrow[\text{}]{\text{graph mining}}  \{r_j^{\ast}\}_{j=1}^l$$

Several approaches in the literature formulate the ensemble consensus function as a graph partitioning problem \cite{strehl2002cluster,Fern04,Domeniconi09}. Our aim here is different, since we are not concerned with the aggregation step. Instead, we organize the components (rankings) in a graph, with the goal of removing poor rankings from further consideration. The challenge is to relate the quality of the rankings with the structure of the graph, in absence of supervision. 

We define two instances of the graph-based ranking selection class, called {\textit{\CORE}} and {\textit{\CROWD}}. In the \CORE approach, we map the problem of selecting ensemble components onto a community detection problem in a graph.
To this end, given the ranking ensemble $\{ r_j\}_{j=1}^m$, we construct a complete weighted graph $G_c=(V, W^c)$, where the vertices in $V$ correspond to the rankings, and $|V|=m$. In connecting the vertices with one another, we want the good rankings to form strongly connected components, so that they can emerge as a dense community. Looking at the rankings in Figure \ref{fig:comp_dist}, we observe that pairs of good rankings are similar in their top ranked objects, while a good and a poor rankings will be dissimilar in the way they rank objects at the top. As such, a similarity measure that emphasizes the top ranked points will in general consider two good rankings as more similar than a good and a poor rankings. The weighted Kendal tau correlation measure \cite{vigna2015weighted} is a measure of similarity that satisfies this property. Let's consider, as an example, Figure \ref{tab:wtau}, which shows the weighted Kendall tau similarity values between all pairs of rankings given in Figure \ref{fig:comp_dist}. We observe high values for all pairs between 1 and 4 (the good rankings), and significantly smaller values for any ranking 1 through 4 and rankings 19 and 20 (the poor rankings).  This suggests the following definition of $W^c$.

%\begin{table}[!htbp]
%  \centering
%    \begin{adjustbox}{width=0.4\columnwidth}
%	\def\arraystretch{2}
%    \begin{tabular}{c  c c c c c c}
%     & \textbf{1} & \textbf{2} & \textbf{3} & \textbf{4} & \textbf{19} & \textbf{20} \\
%    \myline
%    \textbf{1}  & -& 0.92& 0.91 &0.91& 0.55 &0.38 \\
%	\textbf{2}  & 0.92& - &0.93 &0.89 &0.52 &0.38 \\
%	\textbf{3}  & 0.91& 0.93& -& 0.90& 0.53& 0.39 \\
%	\textbf{4}  & 0.91& 0.89& 0.90& -& 0.55& 0.41 \\
%	\textbf{19} & 0.55& 0.52& 0.53& 0.55& -& 0.58 \\
%	\textbf{20} & 0.38& 0.38& 0.39& 0.41& 0.58& - \\
%    \myline
%  \end{tabular}
%  \end{adjustbox}
%  \vspace{1em}
%  \caption{Weighted Kendall tau similarity values between the six rankings given in Figure \ref{fig:comp_dist}. \label{tab:wtau}}  
%\end{table}

For every two distinct rankings  $r_i$ and $r_j$, $W^c_{ij} = K_w(r_i, r_j)$, where $K_w$ is the weighted Kendall tau measure. $W^c_{ii}=0$, for all $i$. In order to enable the strongly connected components to emerge, we then prune the edges in $G_c$ by retaining only the edges corresponding to the largest $m$ $W^c_{ij}$ values, where $m$ is equal to the number of vertices. We observe that, a connected graph with $m$ vertices, has at least $m-1$ edges. In pruning the edges we wanted to enable  graph connectivity, thus the choice of $m$ as threshold on the number of edges. Figure \ref{fig:graph} shows the graph $G$ obtained for the ensemble of 20 rankings of Figure \ref{fig:comp_dist}. Notably, the five best components form a 5-clique in this case, which is also the $k$-core subgraph for the resulting $G$, with the largest $k$ ($k=4$). A $k$-core subgraph is the maximal connected subgraph of $G$ in which all vertices have degree at least $k$. Nodes 19 and 20 (the poorest rankings) end up being isolated nodes.

We also observe that the three poor rankings 15, 16, and 17 in Figure \ref{fig:graph} form a 3-clique, revealing pair-wise correlations superior to the threshold. Under the assumption that the ranking components are affected by {\textit{diverse errors}}, cliques of ``poor'' rankings will stay small, and the $k$-core subgraph, with the largest $k$, can identify the subset of rankings of good quality to provide as input to the aggregation function.
We call this ranking selection algorithm {\textit{\CORE}}, and summarize its steps in Algorithm \ref{alg:RS1}.

The \CROWD ranking selection technique takes a different approach to prune the complete weighted graph $G_c$. The \CORE technique typically retains a minority of the ensemble components (25\% on average in our experiments). 
In an effort of selecting a larger number of (good) components, rather than keeping the components in the largest $k$-core, we discard the ones deemed as poor, and keep the remaining. To estimate the poor components we proceed as follows. For each vertex $v_i$ in $G_c$, we compute its weighted degree $d_i=\sum_{j=1}^m W_{ij}^c$. Under the assumption that rankings make diverse errors, we expect poor components to have small weights associated to the incident edges, and therefore a low weighted degree. For example, considering the adjacency matrix in Figure \ref{tab:wtau}, the weighted degrees of the vertices are: $d_1=3.67$, $d_2=3.64$, $d_3=3.66$, $d_4=3.66$, $d_{19}=2.73$, $d_{20}=2.14$; hence, the poor components (19 and 20) have the lowest weighted degrees. In our experiments we discard 20\% of the total number of vertices with the lowest $d_i$ values. We call the resulting algorithm {\textit{\CROWD}}. \CROWD strikes to preserve a larger pool of components in comparison to \CORE. A summary of the steps is given in Algorithm \ref{alg:RS2}.

Hierarchical versions of both \CORE and \CROWD can also be adopted. One can run \CORE on independent ensembles, and then aggregate all the selected components in a new ensemble, and run \CORE again on it. We can proceed similarly for \CROWD. If poor components are sifted at each level, improvements upon the \CORE (\CROWD) technique are expected. The depth of the hierarchy can be extended beyond two as well. In our experiments, we test the two-level hierarchy, and call the respective techniques ${\mbox{\CORE}}^2$ and ${\mbox{\CROWD}}^2$.

%\section{Ranking Aggregation}
%\label{sec:rank-agg}

\section{Experiments}
\label{sec:exp}

\subsection{Datasets}
\label{sec:exp:datasets}
To evaluate outlier methods, typically, data for classification is used and adapted to the task of anomaly detection. The majority class, or a combination of different large classes, is considered as the inliers. The rest of the data, mostly downsampled, plays the role of outliers. For our experiments, we used  datasets from two publicly available repositories. In particular, Lymphography, Shuttle, SpamBase, Waveform, WDBC, Wilt, and WPBC were generated as described in   \cite{campos2016evaluation}\footnote{Data available at: http://www.dbs.ifi.lmu.de/research/outlier-evaluation/DAMI/}; Ecoli4, Pima, Segment0, Yeast2v4, and Yeast05679v4 were generated as described in \cite{alcala2009keel,alcala2011keel}\footnote{Data available at: http://sci2s.ugr.es/keel/imbalanced.php\#sub2A}. SatImage is  taken from the UCI repository \cite{blake1998uci}: the majority class is used as inliers, and $0.01$ of the rest of the data is subsampled to derive the outliers. 
A summary of the datasets is available in Table \ref{tab:ds-info}. 
%\begin{table}
%  \centering
%  \begin{adjustbox}{max width=0.98\columnwidth}
%   \setlength\tabcolsep{4.0pt} 
%    \begin{tabular}{l c c c c }
%        DataSet &  \textbf{Instances} & \textbf{Attributes} & \textbf{Outliers \%}\\
%    \myline
%\textbf{Ecoli4}&314&8&6\\
%\textbf{Glass}&214&7&4\\
%\textbf{Lymphography}&148&47&4\\
%\textbf{PageBlocks}&5,473&10&10\\
%\textbf{Pima}&510&8&2\\
%\textbf{SatImage}&1,072&37&3\\
%\textbf{Segment0}&2,308&20&14\\
%\textbf{Shuttle}&1,013&9&1\\
%\textbf{SpamBase}&2,528&59&2\\
%\textbf{Stamps}&340&9&9\\
%\textbf{Waveform}&3,433&21&3\\
%\textbf{WDBC}&357&32&3\\
%\textbf{Wilt}&4,839&5&5\\
%\textbf{WPBC}&198&35&23\\
%\textbf{Yeast05679v4}&528&8&10\\
%\textbf{Yeast2v4}&514&8&10\\
%    \myline
%  \end{tabular}
%  \end{adjustbox} 
%  \vspace{1em}
%  \caption{Characteristics of the datasets for outlier detection used in the experiments.  \label{tab:ds-info}}
%\end{table}
%

\begin{table}
 \centering
 \begin{adjustbox}{width=\textwidth}
    \begin{tabular}{@{} cl*{16}c @{}}
        & \rot{\textbf{Ecoli4}} & \rot{\textbf{Glass}} & \rot{\textbf{Lymphography}} & \rot{\textbf{PageBlocks}} 
        & \rot{\textbf{Pima}} & \rot{\textbf{SatImage}} & \rot{\textbf{Segment0}} 
        & \rot{\textbf{Shuttle}} & \rot{\textbf{SpamBase}} & \rot{\textbf{Stamps}}
        & \rot{\textbf{Waveform}} & \rot{\textbf{WDBC}} & \rot{\textbf{Wilt}}
        & \rot{\textbf{WPBC}} & \rot{\textbf{Yeast05679v4}} & \rot{\textbf{Yeast2v4}}\\
         \myline
\textbf{Instances}             &314 & 214 & 148 & 5,473 & 510 & 1,072 & 2,308 & 1,013 & 2,528 & 340 & 3,433 & 357 & 4,839 & 198 & 528 & 514\\
\textbf{Attributes}               &8 & 7 & 47 & 10 & 8 & 37 & 20 & 9 & 59 & 9 & 21 & 32 & 5 & 35 & 8 & 8\\
\textbf{Outliers \%}              &6 & 4 & 4 & 10 & 2 & 3 & 14 & 1 & 2 & 9 & 3 & 3 & 5 & 23 & 10 & 10\\      
        \myline
    \end{tabular}
     \end{adjustbox}
  \caption{Characteristics of the datasets for outlier detection used in the experiments.  \label{tab:ds-info}}
  \vspace{-2em}
\end{table}

\subsection{Ensemble construction}
\label{sec:exp:build}
To evaluate ranking selection algorithms, we first need to construct an ensemble of outlier rankings. We construct homogeneous ensembles, where all components are generated using the  LOF algorithm \cite{lof} as the base detector. In order to generate diverse components,  we perform subsampling.  It has been shown that subsampling can create diverse outlier components, and under specific conditions can improve the overall performance, compared to using the entire data \cite{zimek2013subsampling}. To encourage diversity, for each component, we randomly select the subsampling rate  from the (\%) values $\{10,15,20,25,30\}$. We also select the value of the {\textit{MinPts}} parameter of LOF in the set $\{3,5,7,9\}$. Once the subsample is selected, for each point in the dataset, the nearest neighbors, their distances, and the relative densities required in the LOF algorithm are calculated only with respect to the points in the subsample and using the selected {\textit{MinPts}} value. In our experiments, we fix the ensemble size to $20$.

\subsection{Consensus functions}
\label{sec:exp:consensus}
Various methods exist in the literature to unify outlier scores obtained from different base detectors \cite{gao2006converting,kriegel2011interpreting}, and to merge different rank lists \cite{kemeny1959mathematics,kolde2012robust}.
However, since in our setting each ensemble component is generated by the LOF algorithm, no unification is needed.

Our methods focus on the design of the selection mechanism of the ensemble components. As such, we use simple consensus functions across all approaches being compared, i.e. \textit{Maximum}, \textit{Average}, and \textit{Minimum} functions. The use of more sophisticated aggregation functions is out of scope for the current study, and will be considered in the future.
The \textit{Maximum} function assigns to each point the largest score among those received by the various rankings. \textit{Average} and \textit{Minimum} work accordingly in the same fashion.
\textit{Maximum} and \textit{Average} are among the most commonly used consensus functions to aggregate rankings \cite{lof,loci,lazarevic2005feature}.

\subsection{Methods and Evaluation}
\label{sec:exp:eval}
We compare our techniques against state-of-the-art selective outlier ensemble methods, namely {\textit{SelectV}} and {\textit{SelectH}} (we used the code available from the author's website) \cite{rayana2015less}, and DivE \cite{schubert12} (we used the implementation provided by the authors of \cite{rayana2015less}). 
The code of \CORE and \CROWD is publicly available\footnote{Code available at: https://github.com/HamedSarvari/Graph-Based-Selective-Outlier-Ensembles}.
In our experiments, we also include the baseline that selects all the components (called {\textit{All}}).  Moreover, to assess the effectiveness of the ensemble, we  apply the simple LOF algorithm \cite{lof} on the whole data. 
We ran all methods (\CORE, \CROWD, SelectV, SelectH, DivE, and All) on multiple independent ensembles, and report the average performance of each method. The LOF baseline was also applied the same number of times on each data set, with a random choice of the {\textit{MinPts}} parameter from the set $\{3,5,7,9\}$. 
Performance is measured using the area under the Precision-Recall (PR) curve, namely \textit{average precision} \cite{saito2015precision}. This measure was also used by the authors of {\textit{SelectV}} and {\textit{SelectH}} to assess their methods  \cite{rayana2015less}. We observe that, although the area under ROC curve is widely used to evaluate outlier detection methods \cite{campos2016evaluation}, it has been shown that the Precision-Recall curve is more informative than ROC plots when evaluating imbalanced datasets \cite{saito2015precision}.

%\hl{Although area under ROC curve is one of the most commonly used measures for }evaluating outlier detection methods \cite{campos2016evaluation}, it has been declared that Precision-Recall curve is more informative than ROC plots when evaluating imbalanced binary datasets \cite{saito2015precision}
%We are reporting the performance of different methods as area under the Precision-Recall (PR) curve. This measure has also been used by a recent baseline on selective outlier ensembles. \cite{rayana2015less}

To run the hierarchical version of \CORE and \CROWD (${\mbox{\CORE}}^2$ and ${\mbox{\CROWD}}^2$, respectively), we consider batches of 20 ensembles. For each batch, we run \CORE (\CROWD) on each ensemble; we then assemble the outputs of 20 selections in a new ensemble, and run \CORE (\CROWD) again on it. The process is repeated for multiple independent batches of 20 ensembles, and average performance is reported for each method. We also run a variant in which, for each batch of 20 ensembles, we just assemble the 20 outputs of \CORE (\CROWD) in a new ensemble and directly apply the consensus function. These variants are called \CORE.U and \CROWD.U.

For a fair comparison, we also set up runs of  SelectV, SelectH, DivE, and All, where we enable the techniques to have access to all the ensembles in each batch. That is, we generate a single ensemble of $20 \times 20 = 400$ components from a given batch, and run each competitor method on it. The  techniques in this setting are denoted as SelectV.U, SelectH.U, DivE.U, and All.U, respectively.

\subsection{Results}
\label{sec:exp:results}
Table \ref{tab:perf-first} gives the average performances (areas under the PR curve) of \CORE, \CROWD, DivE, SelectV, SelectH, and All across all datasets and for the three consensus functions.
For WDBC, WPBC, Pima,Yeast05679v4, Ecoli4, Shuttle, and SpamBase averages are computed over 400 independent ensembles. For the remaining datasets averages are computed over 200 independent ensembles.

Table \ref{tab:perf-second} gives the average performances (areas under the PR curve) of $\mbox{\CORE}^2$, $\mbox{\CROWD}^2$, \CORE.U, \CROWD.U, DivE.U, SelectV.U, SelectH.U, and All.U across all datasets and for the three consensus functions.
For WDBC, WPBC, Pima,Yeast05679v4, Ecoli4, Shuttle, and SpamBase averages are computed over 20 batches (of 20 ensembles each). For the remaining datasets, averages are computed over 10 batches.

For both tables, statistical significance is assessed using a one-way ANOVA with a post-hoc Tukey HSD test with a p-value threshold equal to $0.01$. For each dataset, boldface indicates  the technique with the best performance score, and any technique which is {\textit{not}} statistically significantly inferior to it. For each dataset, the best performance score is also underlined.

%For every algorithm and each consensus method, average performance across all ensembles is reported in Table \ref{tab:perf-first}. In order to verify sigificant of differences in performance of various methods, a one-way ANOVA with post-hoc Tukey HSD test is conducted. The best method across each dataset is bold faced and underlined. More than one boldfaced number in a row is an indicator of the fact that differences among those numbers were not statistically significant.

%\begin{itemize}
%\item Table with  the AUC results (average and std + statistical significance) for all data and methods
%\item difficult vs easy data sets: two tables like the above but separate for easy and difficult data sets?
%\item To analyze easy vs difficult cases: plot LOF scores, histogram of w-tau values, and compute spread of the w-tau values 
%\item We can investigate why Select-V and Select-H do not perform well on some cases by looking into the components each selects vs. those selected by our method, and check their accuracy and correlation as measures by w-tau. (See Table 1 in ``Less is More'')
%\item Diversity vs Accuracy analysis of our method(s)
%\item Running times
%\end{itemize}

\subsection{Analysis}
\label{sec:exp:analysis}

Table \ref{tab:perf-first} shows that, out of the 16 datasets, \CORE and \CROWD are ranked among the top performers in 12 datasets; SelectV and SelectH in 6; DivE and All in 9, and LOF in 3. Overall, our selective techniques are superior against the state-of-the-art approaches for selective outlier ensembles (SelectV, SelectH, and DivE), and against All. In particular, Core and Cull give the {\it best performance scores} (underlined values) in 10 datasets; DivE in 1; SelectH and SelectV in 2; All in 3; and LOF in 3. Core and Cull win by a large margin. 

It's known that the All technique, especially when combined with average, is a strong baseline and hard to defeat  \cite{chiang2015anomaly}. Our results confirm this fact. In particular, SelectV and SelectH are not competitive against All on the wide range of problems considered in our experiments. We also observe that DivE often selects all the components, and therefore reduces to All. \CORE and \CROWD emerge as the strongest competitors against All. Single LOF is among the top performers in only three cases; this supports the overall effectiveness of the constructed ensemble. It's interesting to observe that in two out of these three cases (Glass and Segment0), LOF is the only top performer, indicating that the constructed ensemble did not work well for these two problems, regardless of the selective or consensus techniques used. 
%!TEX root = ./OE.tex
%%%%%%%%%%%%%%%%%%%%%%%%%%%%%%%%%
% Performances Tables
%%%%%%%%%%%%%%%%%%%%%%%%%%%%%%%%%

\begin{table*}[t]
  \centering
      \begin{adjustbox}{width=\textwidth}
      \setlength\tabcolsep{4.0pt} 
    \begin{tabular}{l c c c | c c c | c c c | c c c | c c c | c c c | c  }
    \multirow{2}{*}{\backslashbox{DataSet}{Method}}
	& \multicolumn{3}{c|}{\bmath{\CORE}}&\multicolumn{3}{c|}{\bmath{\CROWD}} 
    & \multicolumn{3}{c|}{\bmath{Dive}}
    & \multicolumn{3}{c|}{\bmath{SelectH}} & \multicolumn{3}{c|}{\bmath{SelectV}}
    & \multicolumn{3}{c}{\bmath{All}} & \multicolumn{1}{c}{\bmath{Lof}} \\
    &avg.&max.&min.&avg.&max.&min.&avg.&max.&min.&avg.&max.&min.&avg.&max.&min.&avg.&max.&min.&avg.\\
    \myline
\textbf{Ecoli4} & \textbf{{0.133}} & \textbf{\underline{0.135}} & 0.125 & \textbf{{0.127}} & 0.124 & 0.108 & 0.123 & 0.115 & 0.094 & 0.123 & 0.114 & 0.093 & 0.121 & 0.112 & 0.094 & 0.123 & 0.114 & 0.093 & 0.053\\

\textbf{Glass} & 0.115 & 0.115 & 0.117 & 0.125 & 0.123 & 0.107 & 0.131 & 0.113 & 0.098 & 0.098 & 0.086 & 0.072 & 0.098 & 0.085 & 0.077 & 0.131 & 0.114 & 0.097 & \textbf{\underline{0.155}}\\

\textbf{Lymphography} & 0.311 & 0.294 & 0.321 & 0.585 & 0.379 & \textbf{\underline{0.651}} & 0.567 & 0.368 & 0.627 & 0.291 & 0.266 & 0.32 & 0.322 & 0.288 & 0.331 & 0.348 & 0.3 & 0.365 & 0.475\\

\textbf{PageBlocks} & 0.413 & 0.392 & 0.366 & \textbf{\underline{0.428}} & 0.383 & 0.313 & \textbf{\underline{0.428}} & 0.375 & 0.272 & \textbf{{0.426}} & 0.373 & 0.272 & \textbf{{0.423}} & 0.373 & 0.276 & \textbf{\underline{0.428}} & 0.375 & 0.272 & 0.252\\                                                                                                                                   
\textbf{Pima} & \textbf{\underline{0.028}} & 0.028 & 0.027 & \textbf{{0.027}} & \textbf{\underline{0.028}} & 0.026 & \textbf{{0.027}} & \textbf{{0.027}} & 0.024 & 0.026 & 0.026 & 0.024 & 0.026 & 0.025 & 0.024 & \textbf{{0.027}} & 0.027 & 0.024 & 0.02\\                                                                                                                   
\textbf{SatImage} & \textbf{{0.514}} & \textbf{{0.521}} & 0.472 & 0.505 & \textbf{\underline{0.526}} & 0.36 & 0.479 & \textbf{{0.523}} & 0.231 & 0.464 & 0.502 & 0.23 & 0.46 & 0.5 & 0.232 & 0.486 & \textbf{\underline{0.526}} & 0.232 & 0.21\\                                                                                                                                           
\textbf{Segment0} & 0.104 & 0.105 & 0.108 & 0.103 & 0.106 & 0.111 & 0.103 & 0.108 & 0.113 & 0.104 & 0.108 & 0.113 & 0.104 & 0.109 & 0.113 & 0.103 & 0.108 & 0.113 & \textbf{\underline{0.118}}\\

\textbf{Shuttle} & 0.157 & 0.157 & 0.139 & \textbf{\underline{0.168}} & 0.16 & 0.129 & \textbf{{0.17}} & 0.152 & 0.134 & 0.151 & 0.138 & 0.131 & 0.126 & 0.123 & 0.124 & \textbf{{0.17}} & 0.152 & 0.134 & 0.113\\                                                                                                                                                                                       
\textbf{SpamBase} & 0.088 & 0.092 & 0.076 & 0.091 & 0.095 & 0.068 & 0.094 & 0.096 & 0.064 & 0.126 & \textbf{\underline{0.129}} & 0.076 & 0.125 & \textbf{{0.128}} & 0.079 & 0.094 & 0.096 & 0.064 & 0.072\\                                                                                                                                                                                                          
\textbf{Stamps} & 0.081 & 0.077 & 0.088 & 0.087 & 0.077 & \textbf{{0.104}} & 0.089 & 0.072 & \textbf{{0.109}} & 0.076 & 0.061 & 0.093 & 0.077 & 0.062 & 0.089 & 0.089 & 0.073 & \textbf{\underline{0.11}} & 0.098\\                                                                                                                                                                                      
\textbf{Waveform} & 0.115 & \textbf{\underline{0.13}} & 0.098 & 0.105 & 0.123 & 0.081 & 0.099 & 0.115 & 0.071 & 0.105 & 0.117 & 0.082 & 0.101 & 0.117 & 0.073 & 0.099 & 0.115 & 0.071 & 0.062\\

\textbf{WDBC} & \textbf{\underline{0.815}} & \textbf{{0.8}} & 0.798 & \textbf{\underline{0.815}} & \textbf{{0.8}} & 0.798 & \textbf{{0.814}} & 0.796 & 0.747 & 0.752 & 0.732 & 0.688 & 0.749 & 0.735 & 0.716 & \textbf{{0.814}} & 0.796 & 0.748 & 0.514\\                                                                                                                      
\textbf{Wilt} & 0.075 & 0.063 & \textbf{{0.084}} & 0.076 & 0.059 & \textbf{\underline{0.085}} & 0.078 & 0.058 & \textbf{{0.083}} & 0.076 & 0.058 & \textbf{{0.083}} & 0.078 & 0.058 & \textbf{{0.084}} & 0.078 & 0.058 & \textbf{{0.084}} & 0.071\\                                                                                                                  
\textbf{WPBC} & \textbf{\underline{0.226}} & \textbf{\underline{0.226}} & \textbf{{0.224}} & \textbf{{0.225}} & \textbf{{0.225}} & 0.222 & \textbf{{0.224}} & \textbf{{0.225}} & 0.219 & 0.224 & \textbf{{0.224}} & 0.222 & 0.223 & \textbf{{0.224}} & 0.219 & \textbf{{0.224}} & \textbf{{0.225}} & 0.219 & 0.21\\

\textbf{Yeast05679v4} & 0.132 & 0.131 & 0.132 & 0.133 & 0.132 & \textbf{{0.134}} & \textbf{{0.134}} & 0.133 & \textbf{{0.136}} & 0.133 & 0.132 & \textbf{{0.136}} & 0.133 & 0.132 & \textbf{{0.135}} & \textbf{{0.134}} & 0.133 & \textbf{{0.136}} & \textbf{\underline{0.137}}\\                                                            
\textbf{Yeast2v4} & 0.207 & 0.22 & 0.187 & 0.19 & 0.219 & 0.158 & 0.186 & 0.206 & 0.153 & 0.228 & \textbf{\underline{0.239}} & 0.188 & 0.224 & \textbf{{0.234}} & 0.188 & 0.186 & 0.208 & 0.154 & 0.147\\
\myline
  \end{tabular}
  \end{adjustbox}
  \caption{Average performance (area under the PR curve) for all methods and datasets (no hierarchy). \label{tab:perf-first}}
  \vspace{-2em}
\end{table*}

\begin{table*}[t]
  \centering
      \begin{adjustbox}{width=\textwidth}
      \setlength\tabcolsep{4.0pt} 
    \begin{tabular}{l c c c | c c c | c c c | c c c }

\multirow{2}{*}{\backslashbox{DataSet}{Method}} &  \multicolumn{3}{c|}{\bmath{\CORE^2}} & \multicolumn{3}{c|}{\bmath{\CROWD^2}} & \multicolumn{3}{c|}{\bmath{\CORE.U}} & \multicolumn{3}{c}{\bmath{\CROWD.U}} \\
          & avg. & max. & min. & avg. & max. & min. & avg. & max. & min. & avg. & max. & min. \\
             \myline
\textbf{Ecoli4} & \textbf{{0.14}} & \textbf{\underline{0.141}} & \textbf{{0.131}} & \textbf{{0.13}} & \textbf{{0.131}} & \textbf{{0.116}} & \textbf{{0.135}} & \textbf{{0.139}} & \textbf{{0.128}} & \textbf{{0.129}} & \textbf{{0.123}} & 0.089\\
\textbf{Glass} & \textbf{{0.109}} & \textbf{{0.115}} & \textbf{{0.116}} & \textbf{{0.119}} & \textbf{\underline{0.136}} & \textbf{{0.102}} & \textbf{{0.117}} & \textbf{{0.105}} & \textbf{{0.118}} & \textbf{{0.126}} & \textbf{{0.111}} & 0.076\\
\textbf{Lymphography} & 0.28 & 0.269 & 0.289 & 0.527 & 0.282 &  \textbf{\underline{0.692}} & 0.31 & 0.27 & 0.336 &  \textbf{{0.621}} & 0.293 &  \textbf{{0.69}}\\
\textbf{PageBlocks} & \textbf{{0.416}} & 0.391 &  \textbf{{0.399}} & \textbf{{0.437}} & 0.356 & 0.303 &  \textbf{{0.436}} & 0.368 & 0.323 &  \textbf{\underline{0.439}} & 0.354 & 0.261\\
\textbf{Pima} & \textbf{{0.03}} & \textbf{\underline{0.032}} & \textbf{{0.03}} & 0.027 & 0.029 & 0.025 & 0.028 &  \textbf{{0.03}} & 0.026 & 0.028 & 0.028 & 0.022\\
\textbf{SatImage} & \textbf{{0.524}} & \textbf{{0.532}} & \textbf{{0.486}} & \textbf{{0.519}} & \textbf{{0.535}} & 0.332 &  \textbf{{0.522}} & \textbf{{0.535}} & 0.413 &  \textbf{{0.514}} & \textbf{\underline{0.551}} & 0.224\\
\textbf{Segment0} & 0.103 & 0.104 &  \textbf{{0.108}} & 0.102 & 0.103 &  \textbf{{0.11}} & 0.102 & 0.102 &  \textbf{{0.11}} & 0.102 & 0.104 &  \textbf{{0.114}}\\
\textbf{Shuttle} & \textbf{{0.158}} & \textbf{{0.158}} & 0.141 &  \textbf{{0.17}} & \textbf{{0.171}} & 0.138 &  \textbf{{0.165}} & \textbf{{0.172}} & 0.143 &  \textbf{{0.172}} & \textbf{{0.158}} & 0.135\\
\textbf{SpamBase} & 0.084 & 0.096 & 0.071 & 0.09 & 0.098 & 0.059 & 0.089 & 0.094 & 0.064 & 0.091 & 0.106 & 0.055\\
\textbf{Stamps} & 0.071 & 0.076 & 0.088 & 0.08 & 0.076 &  \textbf{{0.1}} & 0.077 & 0.068 &  \textbf{{0.108}} & 0.093 & 0.078 &  \textbf{{0.119}}\\
\textbf{Waveform} & 0.108 & 0.121 & 0.096 & 0.11 &  \textbf{{0.146}} & 0.079 & 0.114 &  \textbf{\underline{0.159}} & 0.087 & 0.105 & 0.13 & 0.067\\
\textbf{WDBC} & \textbf{\underline{0.815}} & \textbf{{0.806}} & \textbf{{0.803}} & \textbf{\underline{0.815}} & \textbf{{0.806}} & \textbf{{0.803}} & \textbf{\underline{0.815}} & \textbf{{0.798}} & \textbf{{0.768}} & \textbf{\underline{0.815}} & \textbf{{0.798}} & \textbf{{0.768}}\\
\textbf{Wilt} & \textbf{{0.079}} & 0.065 &  \textbf{{0.091}} & \textbf{{0.077}} & 0.054 &  \textbf{{0.089}} & \textbf{{0.075}} & 0.053 &  \textbf{\underline{0.092}} & \textbf{{0.077}} & 0.053 &  \textbf{{0.089}}\\
\textbf{WPBC} & \textbf{{0.23}} & \textbf{\underline{0.232}} & \textbf{{0.228}} & \textbf{{0.226}} & 0.224 & 0.224 &  \textbf{{0.227}} & \textbf{{0.227}} & \textbf{{0.225}} & \textbf{{0.225}} & \textbf{{0.228}} & 0.224\\
\textbf{Yeast05679v4} & \textbf{{0.131}} & \textbf{{0.131}} & \textbf{{0.132}} & \textbf{{0.132}} & \textbf{{0.133}} & \textbf{{0.131}} & \textbf{{0.132}} & \textbf{{0.133}} & \textbf{{0.132}} & \textbf{{0.133}} & \textbf{{0.136}} & \textbf{{0.13}}\\
\textbf{Yeast2v4} & \textbf{{0.255}} & \textbf{\underline{0.268}} & \textbf{{0.246}} & 0.195 &  \textbf{{0.235}} & 0.157 & 0.207 &  \textbf{{0.241}} & 0.171 & 0.191 &  \textbf{{0.22}} & 0.147\\
            \myline
 \multirow{2}{*}{\backslashbox{DataSet}{Method}} & \multicolumn{3}{c|}{\bmath{Dive.U}} & \multicolumn{3}{c|}{\bmath{SelectH.U}} & \multicolumn{3}{c|}{\bmath{SelectV.U}} & \multicolumn{3}{c}{\bmath{All.U}} \\
          & avg. & max. & min. & avg. & max. & min. & avg. & max. & min. & avg. & max. & min. \\
                      \myline
\textbf{Ecoli4} & 0.097 & 0.097 & 0.097 &  \textbf{{0.125}} & 0.107 & 0.076 &  \textbf{{0.126}} & \textbf{{0.115}} & 0.08 &  \textbf{{0.125}} & 0.107 &  0.076\\
\textbf{Glass} & 0.106 & 0.106 & 0.106 & 0.1 & 0.077 & 0.053 & 0.1 & 0.083 & 0.052 &  \textbf{{0.134}} & 0.094 &  0.07\\                                                                        
\textbf{Lymphography} & 0.444 & 0.444 & 0.444 & 0.288 & 0.256 & 0.336 & 0.325 & 0.261 & 0.314 & 0.353 & 0.277 &  0.362\\                                                                                                                                                                                                                                                                                     
\textbf{PageBlocks} & 0.318 & 0.318 & 0.318 &  \textbf{{0.435}} & 0.32 & 0.214 &  \textbf{{0.432}} & 0.327 & 0.217 &  \textbf{{0.437}} & 0.322 &  0.214\\                                                                                                                                                           
\textbf{Pima} & 0.024 & 0.024 & 0.024 & 0.027 & 0.026 & 0.02 & 0.027 & 0.026 & 0.022 & 0.028 & 0.027 &  0.02\\                                                                                                                                                                                                                                                                 
\textbf{SatImage} & 0.429 & 0.429 & 0.429 &  \textbf{{0.469}} & \textbf{{0.526}} & 0.156 & 0.453 &  \textbf{{0.508}} & 0.176 &  \textbf{{0.492}} & \textbf{\underline{0.551}} & 0.156\\                  
\textbf{Segment0} & \textbf{\underline{0.115}} & \textbf{\underline{0.115}} & \textbf{\underline{0.115}} & 0.102 &  \textbf{{0.108}} & \textbf{{0.112}} & 0.103 &  \textbf{{0.108}} & \textbf{{0.114}} & 0.102 &  \textbf{{0.108}} & \textbf{{0.112}}\\                                           
\textbf{Shuttle} & 0.118 & 0.118 & 0.118 &  \textbf{{0.157}} & 0.127 & 0.135 & 0.124 & 0.123 & 0.122 &  \textbf{\underline{0.173}} & 0.135 &  0.143\\                                                                                                                
\textbf{SpamBase} & 0.081 & 0.081 & 0.081 &  \textbf{{0.132}} & \textbf{\underline{0.138}} & 0.057 &  \textbf{{0.132}} & \textbf{{0.131}} & 0.059 & 0.095 & 0.113 &  0.05\\                                                                                                                                                                                                                                                          
\textbf{Stamps} & 0.079 & 0.079 & 0.079 & 0.085 & 0.055 &  \textbf{{0.114}} & 0.082 & 0.057 &  \textbf{{0.099}} & \textbf{{0.099}} & 0.066 &  \textbf{\underline{0.137}}\\                                                                                                                                                                                        
\textbf{Waveform} & 0.093 & 0.093 & 0.093 & 0.11 &  \textbf{{0.141}} & 0.069 & 0.1 &  \textbf{{0.14}} & 0.063 & 0.099 &  \textbf{{0.136}} & 0.06\\
\textbf{WDBC} & \textbf{{0.773}} & \textbf{{0.773}} & \textbf{{0.773}} & 0.755 & 0.726 & 0.655 & 0.751 & 0.737 & 0.708 &  \textbf{\underline{0.815}} & \textbf{{0.791}} & 0.731\\                                                                                                  
\textbf{Wilt} & 0.072 & 0.065 &  \textbf{{0.077}} & \textbf{{0.078}} & 0.051 &  \textbf{{0.089}} & \textbf{{0.079}} & 0.052 &  \textbf{{0.09}} & \textbf{{0.079}} & 0.051 &  \textbf{{0.09}}\\ 
\textbf{WPBC} & 0.219 &  \textbf{{0.226}} & 0.203 &  \textbf{{0.226}} & \textbf{{0.227}} & 0.223 & 0.224 &  \textbf{{0.228}} & 0.218 &  \textbf{{0.225}} & 0.224 &  0.216\\                       
\textbf{Yeast05679v4} & 0.126 & 0.126 & 0.126 &  \textbf{{0.133}} & \textbf{{0.136}} & \textbf{{0.133}} & \textbf{{0.132}} & \textbf{\underline{0.137}} & \textbf{{0.131}} & \textbf{{0.134}} & \textbf{\underline{0.137}} & \textbf{{0.133}}\\
\textbf{Yeast2v4} & 0.186 & 0.186 & 0.186 &  \textbf{{0.233}} & \textbf{{0.221}} & 0.161 &  \textbf{{0.232}} & \textbf{{0.234}} & 0.166 & 0.187 & 0.197 &  0.137\\

\myline
  \end{tabular}
  \end{adjustbox}
  \caption{Average performance (area under the PR curve) for all methods and datasets (with hierarchy). \label{tab:perf-second}}
  \vspace{-2em}
\end{table*}

The results reveal an interesting  fact about \CORE and \CROWD: they manifest their best behavior under different scenarios. They are among the best performing methods in 7 and 11 datasets, respectively, of which only 6 are in common.  On the other hand,  SelectH and SelectV seem highly correlated. They both become competitive in the same 6 datasets.
 The complementary nature of \CORE and \CROWD enables them to succeed in a wide range of problems, since they seem to induce a different learning bias. This opens a new research path to investigate characteristics of the datasets to which  each of these methods is tuned, and suggests the potential for a hybrid approach that leverages their diversity.
 
Table  \ref{tab:perf-second} shows that $\mbox{\CORE}^2$ and $\mbox{\CROWD}^2$ are ranked among the top performers in 15 datasets; Core.U and Cull.U in 15 as well; DivE.U in 4; SelectV.U and SelectH.U in 12; and All.U in 12. Again, the hierarchical versions of our techniques emerge as the strongest competitors. 
In particular, $\mbox{\CORE}^2$ and $\mbox{\CROWD}^2$ give the {\it best performance scores} (underlined values) in 7 datasets; Core.U and Cull.U in 5; DivE.U in 1; SelectH.U and SelectV.U in 2; and All.U in 5.

The two-level pruning mechanisms of $\mbox{\CORE}^2$ and $\mbox{\CROWD}^2$  effectively prune poor components among a large pool of rankings. On the other hand, All.U deals with large ensembles, which are likely to contain a fair number of poor components, thus hurting the relative performance against the competitors. Overall, the behavior of SelectV.U and SelectH.U is comparable to All.U, while DivE.U gives the worst performance.  
%These results are significant, since it's known that the All technique (especially when combined with average) is a strong baseline and hard to defeat  \cite{chiang2015anomaly}. 

An insightful observation from the results in Table  \ref{tab:perf-second} is the strong performance of Core.U.  It's among the best performers in 14 (out of 16) datasets, and its overall performance is superior to $\mbox{\CORE}^2$. In a way, Core.U achieves the best-of-both-worlds: it  first uses Core to discard poor components across different ensembles; then it aggregates all selected rankings, acting like All, but on a "boosted" pool of components. 
$\mbox{\CORE}$.U is superior to (or tied with) All.U in almost all scenarios (15 out of 16), and therefore a very promising candidate for outlier ensemble selection.

We finally observe that the best performing consensus functions depend on the dataset, and to a less extent on the method. A deeper understanding of this behavior is in our agenda for future work.
%The selected components available after applying $\mbox{\CORE}$ once are already purified and preserving all of them give more boost to ensemble than removing some. On the other hand All.U uses all ensembles with no filtering mechanism which degrades the overal quality.
%In other words $\mbox{\CORE}$.U is benefiting both from selection strategy of $\mbox{\CORE}$ and inclusiveness of All.

%$\mbox{\CORE}$.U is superior to (or tied with) All.U in almost all scenarios (15 out of 16), and therefore a very promising candidate for outlier ensemble selection. $\mbox{\CORE}$.U can obviously beat All.U is a very though baseline and also state-of-the-art selective methods, Select V and Select H, which can be competitive in at most 7 datasets in hierarchical scenario.

\section{Complexity Analysis}
\label{sec:exp:analysis:complexity}
We analyze the theoretical complexity for all the considered methods. For simplicity, we omit the cost of sampling, the cost needed to compute the anomaly scores,  and the cost of running the aggregation function. These steps are common to all methods. 
Let $n$ be the size of each ranking and $m$ the size of the ensemble.
\begin{itemize}
\item \textit{All}:  The cost of selecting all the components is simply equal to the size of the ensemble: $\mathbf{m}$.

\item \textit{\CORE and \CROWD}:  The cost of the graph construction is $m \cdot m \cdot n \cdot log (n)$, obtained by multiplying the number of edges with the cost of computing the weighted tau, which is $n \cdot log (n)$  as reported in \cite{vigna2015weighted}. The rest of the computation is linear in the number of edges, which is $m \cdot m$. So the total cost is:
$ (\mathbf{m \cdot m \cdot n \cdot log (n)}) + (m \cdot m)$,
which is dominated by the factor $\mathbf{n \cdot log (n) \cdot m^2}$.

\item \textit{DivE}: As reported in \cite{schubert12}, the first operation performed by DivE is the \textit{Union} of the top-\textit{k} outliers, with a  cost of $n\cdot m$ when $k=n$. 
The next step consists in sorting the converted rankings using the weighted Pearson correlation, with a cost of $m\cdot n +  m\cdot log(m)$. This includes the  cost of $m$ Pearson's coefficients ($m\cdot n$), and the cost of sorting the $m$ rankings according to the elaborated coefficients. 
The computation of sorting the converted rankings using the weighted Pearson correlation is repeated two times before the loop that contain the $m$ rankings. As a result,  the overall cost of DivE is:
$ n\cdot m  + (\mathbf{(m+2)\cdot(m\cdot n +  m\cdot log(m))})$,
which it is dominated by the factor $\mathbf{n \cdot m^2}$.

\item \textit{Select-V}: As reported in \cite{rayana2015less}, the first operation performed by SelectV is \textit{Unification}, which converts scores to probability estimates. Even if we consider as constant the cost of  Unification, the total cost of performing Unification for all rankings is $n\cdot m$. The cost of rank sorting is $n\cdot log(n)$.
The next step consists in sorting the converted rankings using the weighted Pearson correlation, with a cost of $m\cdot n +  m\cdot log(m)$, exactly as in DivE. 
%This includes the computation cost of $m$ Pearson's coefficients ($m\cdot n$) and the cost of sorting the $m$ rankings according to the elaborated coefficients. 
The computation of sorting the converted rankings using the weighted Pearson correlation is repeated $m+1$ times, and the final running cost to perform SelectV is:
$ n\cdot m + n\cdot log(n) + (\mathbf{(m+1)\cdot(m\cdot n +  m\cdot log(m))})$,
which it is dominated by the factor $\mathbf{n \cdot m^2}$.

\item \textit{Select-H}: As reported in \cite{rayana2015less}, the first expensive procedure performed is the computation of \textit{MixtureModel}. Its cost depends on the number of iterations $i$, which was set to 100 as suggested in \cite{rayana2015less}, on the size $m$ of the ensemble, and on the length $n$ of the score vectors; the resulting cost is $i\cdot n \cdot m$. The second expensive procedure  is \textit{RobustRankAggregation}, which costs $n \cdot m$.
The subsequent loop  is dominated by the number of estimated outliers, and in the worst case its cost is $n^2 \cdot log(n)$. The algorithm concludes with a clustering phase (k-means \cite{Lloyd:2006:LSQ:2263356.2269955} is the used algorithm), which costs $2 \cdot m\cdot n$. The final running cost to perform SelectH is:
$ (i\cdot n \cdot m) + (n \cdot m) + (\mathbf{n^2 \cdot log(n)})+ (2 \cdot m\cdot n)$,
which is dominated by the factor $\mathbf{n^2 \cdot log(n)}$.
\end{itemize}

The theoretical analysis shows that SelectH is dominated by $n^2$, our methods \CORE and \CROWD by $n\cdot log(n)$, and DivE and SelectV by $n$. In real world scenarios, as the number of data  grows large,  SelectH may become prohibitively expensive.

\begin{table}
  \centering
    \begin{adjustbox}{width=1 \columnwidth}
	\begin{tabular}{r c c c c | c c c c}

&\rotqb{\textbf{Core/Cull}}&\rotqb{\textbf{Dive }}&\rotqb{\textbf{SelectH}}&\rotqb{\textbf{SelectV}}&\rotqb{\textbf{Core-U/Cull-U}}&\rotqb{\textbf{DiveU}}&\rotqb{\textbf{SelectHU}}&\rotqb{\textbf{SelectVU}}\\
        \myline
\textbf{Ecoli4} & 0.1955 & 0.0407 & 0.3507 & 0.0139 & 3.7343 & 2.4337 & 9833.1550 & 1.8934\\
\textbf{Glass} & 0.1202 & 0.0310 & 0.2313 & 0.0124 & 5.1370 & 2.2313 & 6139.0900 & 1.6363\\
\textbf{Lymphography} & 0.0918 & 0.0228 & 0.2686 & 0.0107 & 37.3748 & 2.1586 & 1798.3125 & 0.7797\\
\textbf{PageBlocks} & 3.5285 & 0.1021 & 7.7770 & 0.1160 & 98.0259 & 18.2131 & 134998.5714 & 11.9919\\
\textbf{Pima} & 0.3028 & 0.0286 & 0.9487 & 0.0205 & 5.3593 & 2.8979 & 14137.0000 & 2.5817\\
\textbf{SatImage} & 0.7446 & 0.0323 & 1.2203 & 0.0353 & 7.4759 & 4.0789 & 30714.4000 & 3.4625\\
\textbf{Segment0} & 1.5537 & 0.0691 & 3.3306 & 0.0481 & 20.7469 & 6.0357 & 64577.3000 & 5.9603\\
\textbf{Shuttle} & 0.6247 & 0.0346 & 1.1076 & 0.0214 & 9.6138 & 3.9340 & 27573.0500 & 3.0115\\
\textbf{SpamBase} & 1.9090 & 0.0456 & 4.8094 & 0.0448 & 14.6854 & 7.1720 & 71212.8333 & 6.4726\\
\textbf{Stamps} & 0.1834 & 0.0265 & 0.3725 & 0.0136 & 1.5201 & 2.3445 & 9135.5000 & 1.8553\\
\textbf{Waveform} & 2.3860 & 0.0553 & 6.5074 & 0.0693 & 73.8695 & 9.1355 & 103188.4444 & 9.5774\\
\textbf{WDBC} & 0.2162 & 0.0330 & 0.3878 & 0.0138 & 1.4657 & 2.5056 & 10659.8000 & 1.8951\\
\textbf{Wilt} & 3.3298 & 0.0648 & 4.6988 & 0.0887 & 29.2481 & 13.7465 & 117418.8889 & 10.7646\\
\textbf{WPBC} & 0.1116 & 0.0647 & 1.9563 & 0.0175 & 1.7298 & 2.6575 & 3197.7000 & 1.0396\\
\textbf{Yeast2v4} & 0.3031 & 0.0373 & 0.5683 & 0.0152 & 3.7115 & 2.6427 & 14770.1000 & 2.2619\\
\textbf{Yeast05679v4} & 0.3110 & 0.0538 & 0.5941 & 0.0154 & 3.8485 & 2.8501 & 15323.2500 & 2.3146\\
\hline
Average & 0.2532 & 0.0473 & 0.4724 & 0.0146 & 3.7914 & 2.6419 & 12578.2025 & 2.1040\\
\myline
  \end{tabular}
  \end{adjustbox}  
  \caption{Running times of experiments in Table \ref{tab:perf-second}  expressed in seconds. \label{tab:complexity}} 
  \vspace{-2em}
\end{table}

%\subsection*{Running times}

We have also computed the empirical running times. For each dataset, the average running time of all the runs for each method is recorded  in Table \ref{tab:complexity}. Experiments were run on a laptop with an Intel Core i7 Processor @2.80GHz and 16GB RAM. The empirical running times are consistent with the complexity analysis given above.
In particular, we observe that the running times of DivE and SelectV have the same order of magnitude. Core and Cull are faster then SelectH but slower than DivE and SelectV.
When the number of components increases, the running times of Core.U, Cull.U, Dive.U, and SelectV.U are almost identical;
in contrast, the runnning time of SelectH.U increases much more rapidly with respect to the other methods.

\section{Conclusion}
\label{sec:conclusion}
We have introduced a new graph-based class of ranking selection methods for outlier ensembles. In particular, we have defined two specific approaches, \CORE and \CROWD, and hierarchical extensions of the same. Our extensive evaluation on a variety of heterogeneous data and methods shows that our approach outperforms state-of-the-art selective outlier ensemble techniques in a number of cases. Interesting and challenging questions are open for future investigation, including a characterization of the scenarios when \CORE outperforms \CROWD, or viceversa; studying how our selective techniques affect the accuracy/diversity tradeoffs; exploring hybrid methods, different outlier detection techniques and alternative consensus functions, and analyze their effects in more depth.

\bibliography{References}
\bibliographystyle{ieeetr}

\end{document}